\documentclass{article}
\usepackage{spconf,amsmath,graphicx}
\usepackage{enumitem}
\usepackage{subcaption}

\title{
Live Video Synopsis for Multiple Cameras
}

\name{Yedid Hoshen ~~~~~~~~~~~  Shmuel Peleg}
\address{School of Computer Science and Engineering\\
        The Hebrew University of Jerusalem, Israel}

\begin{document}

\maketitle

\begin{abstract}
Video surveillance cameras generate most of recorded video, and there is far more recorded video than operators can watch. Much progress has recently been made using summarization of recorded video, but such techniques do not have much impact on live video surveillance.

We assume a camera hierarchy where a Master camera observes the decision-critical region, and one or more Slave cameras observe regions where past activity is important for making the current decision. We propose that when people appear in the live Master camera, the Slave cameras will display their past activities, and the operator could use past information for real-time decision making. 

The basic units of our method are action tubes, representing objects and their trajectories over time. Our object-based method has advantages over frame based methods, as it can handle multiple people, multiple activities for each person, and can address re-identification uncertainty.

\end{abstract}


\vspace{-0.1cm}
\section{Introduction}
\label{sec:intro}
\vspace{-0.1cm}

Surveillance cameras are installed everywhere, and are becoming even more popular due to lower costs of cameras, networking, and storage. The increase in the number of cameras is not being offset by a proportional increase in the number of operators available to monitor the video, and in practice most surveillance video is not being viewed. The large gap between the availability of human operators and the need to extract the information in the recorded video has attracted much interest from the computer vision community.

Surveillance video has two main purposes: real-time remote sensing and forensic historical analysis, where historical video is rarely used for real-time decision making. In this paper we suggest a novel object-based method for using past surveillance video for live decision making.  

One of the great challenges for human operators is being able to exploit relations between the video streams across time and across cameras. Let us consider a library with several cameras. Some cameras view the bookshelves while one camera views the lending desk. Viewing each stream independently may not reveal suspicious behavior. The librarian can not remember all activity of all library visitors, which occur at different times in different cameras. However, if all the bookshelf cameras delayed showing the activities of each reader until he reaches the lending desk, the librarian can easily grasp all the reader's activity before he leaves the library.

The camera synchronization paradigm is quite general. Other cases where cross camera relationship is important are:
\begin{itemize}[noitemsep,topsep=2pt,parsep=0pt,partopsep=0pt,leftmargin=*]
\item Effective Business intelligence: What items did customers look at before purchasing? 
\item Anomalous behavior detection: Does a traveler change his pace before going through customs? 
\item Checkpoint: A guard at the exit from a secure facility can observe if visitors behaved suspiciously during their visit before being allowed to leave. 

\end{itemize}

The relations in these cases are all object-based, which benefit from comparing the behavior of people across different locations and times. All such systems are hierarchical, where one camera is viewed in real-time (Master) while other cameras are of forensic significance (Slaves). The Master camera need not be static, and could even be a body mounted camera worn by the operator. In this case the Master video need not be viewed, as its view is the same as the operator's.  

In standard camera networks the analyst needs to remember all objects during a few hours of video, which is unreasonable. However, in a hierarchical camera system, if we display in the Slave videos the previous actions of all persons currently observed in the Master camera, the operator will need to remember only a few seconds of video from the Slave cameras for understanding the activity in the Master camera. This motivates Live Video Synopsis (LVS). In LVS activity tubes are initially extracted from all Slave video streams, and persons are identified and labeled. Tubes are then shifted in time in the Slave videos to be displayed only when the person is observed in the Master view.

Notably, we shift Slave tubes in time but do not attempt to bring tubes from different cameras onto the same screen or shift tubes spatially as object tubes might be placed on semantically unrelated backgrounds sometimes with absurd results (e.g. people floating in mid-air). Also changes in geometry between cameras can cause some tubes to look unnatural and out of place (e.g. front and side views).
	
Live Video Synopsis has the following benefits: 1) The relations between persons observed at multiple cameras are clearly visible to the operator. 2) Multiple persons and histories can be observed on the same screen. 3) In cases of re-identification uncertainty, multiple possibilities can be displayed. 4) The information can aid live decision making.

\vspace{-0.1cm}
\section{Related Work}
\label{sec:related}
\vspace{-0.1cm}

Much work has been done on understanding surveillance video. Popular approaches include the classification of activity as normal/anomalous \cite{jiang2011anomalous,zhao2011online}, or using activity recognition to transcribe surveillance video into words \cite{kojima2002natural,rohrbach13iccv}. High-level activity understanding is a very promising research direction, but current performance has room for improvements. Realizing that the need for human inspection of video will remain for some time, many methods create visual summaries for faster viewing. 

\begin{figure}[tb]
	\centering
		\begin{subfigure}[t]{0.23\textwidth}
                \centering
                \includegraphics[width=1.0\textwidth]{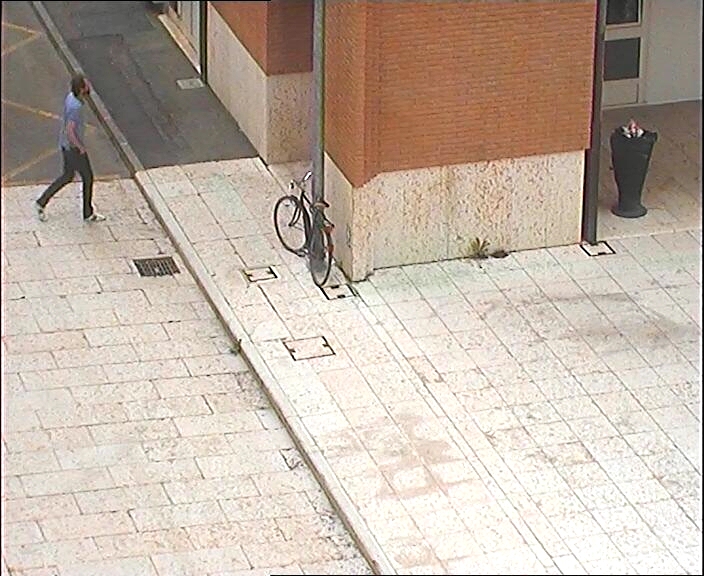}
                \caption{}
                \label{fig:SnapIll1}
    \end{subfigure}
		\begin{subfigure}[t]{0.23\textwidth}
                \centering
                \includegraphics[width=1.0\textwidth]{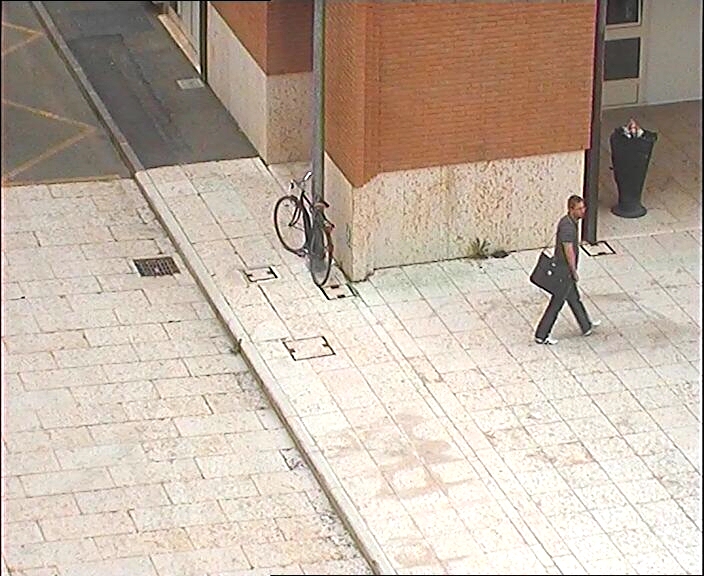}
                \caption{}
                \label{fig:SnapIll2}
    \end{subfigure}
\\
		\begin{subfigure}[t]{0.23\textwidth}
               \centering
                \includegraphics[width=1.0\textwidth]{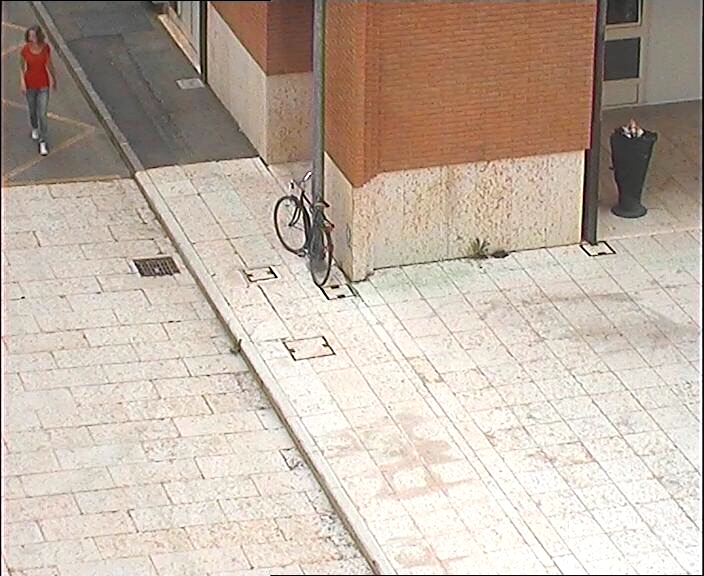}
                \caption{}
                \label{fig:SnapIll3}
    \end{subfigure}
		\begin{subfigure}[t]{0.23\textwidth}
                \centering
                \includegraphics[width=1.0\textwidth]{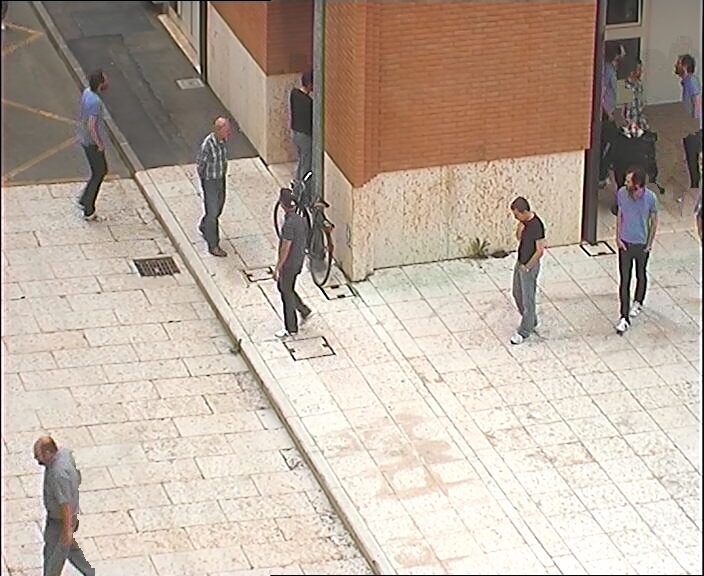}
                \caption{}
                \label{fig:SnapIll4}
    \end{subfigure}
	\caption{Video Synopsis: (a-c) Original frames. (d) Synopsis frame. Objects from different times appear simultaneously.}
	\label{fig:SnapshotIll}
\end{figure}

One approach for visual summarization is the generation of a storyboard by selecting some key frames \cite{gong2000video,khoslalarge}. Another approach is adaptive fast forward \cite{petrovic2005adaptive}, dropping frames at different rates depending on how interesting the video is. Video synopsis \cite{rav2006making,pritch2007webcam,pritch2008nonchronological,feng2012online}, shifting activities in time so that as many activities can be presented simultaneously, presents all activities of a video in a much shorter video. See Fig~\ref{fig:SnapshotIll}.

Single camera approaches for summarization do not generalize well to multiple cameras, as they do not take into account the relationship between the different cameras. Some work addressed video captured by several overlapping cameras \cite{fu2010multi}. But this work can not be used with most cameras which are mostly non-overlapping.

Representation of the video from non-overlapping cameras has received little attention, a notable exception is \cite{sawhney2002video}, which projects multiple video cameras on a 3D model of the environment. But such a 3D model is not generally available. Another interesting work has been done by \cite{porikli2004multi}, who have recognized the importance of using objects for highlighting relationships between video streams from multiple cameras. Their work however has concentrated on the extraction and indexing of objects rather than on visual representation.

A somehow related approach is Multi-Video Browsing and Summarization \cite{dale2012multi}, which attempts to synchronize video streams by shifting frames in time, so that visually similar frames are observed in all videos at the same time.  This scheme measures similarity by a set of trained visual similarity descriptors among frames, in contrast to our work which is object based. 

\vspace{-0.1cm}
\section{Live Video Synopsis (LVS)}
\vspace{-0.1cm}

The generation of LVS consists of three stages: Preprocessing (Sec.~\ref{subsec:pre}), Optimization (Sec.~\ref{subsec:opt}), and Display (Sec.~\ref{subsec:disp}).

\vspace{-0.1cm}
\subsection{Video Preprocessing}
\label{subsec:pre}

\begin{figure}[tb]
	\centering
  \includegraphics[width=0.45\textwidth]{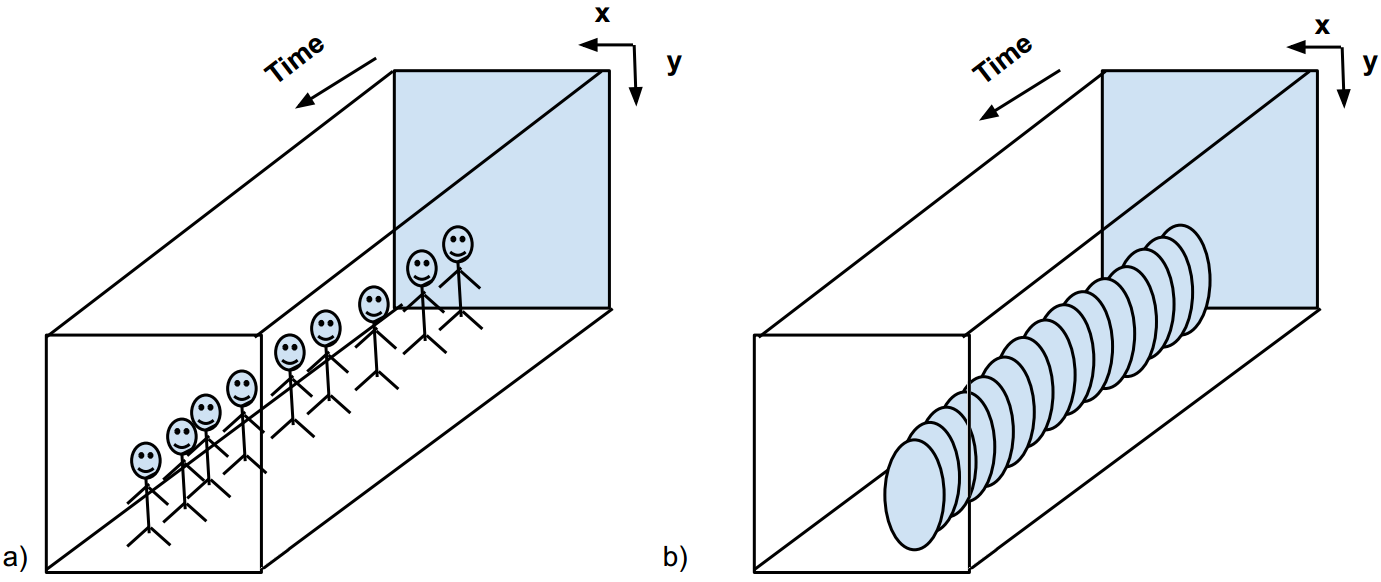}
	\caption{ a) A video showing a single object. b) Tubes are binary masks representing an object, containing all pixels in all frames belonging to the object.}
	\label{fig:TubeEx}
\end{figure}

Before selecting the Slave action tubes corresponding to the persons observed by the Master camera, several pre-processing steps are required:

\begin{enumerate}[noitemsep,topsep=2pt,parsep=0pt,partopsep=0pt,leftmargin=*]

\item People are detected and tracked in all slave video streams. Each person is represented as a space-time ``tube'', which is the union of all pixels of this object in each frame. Relevant literature on object detection using background subtraction appears in \cite{stauffer1999adaptive,ko2008background}, and tracking objects across frames appears in \cite{pritch2008nonchronological,porikli2004multi}. The extraction of video tubes is depicted in Fig.~\ref{fig:TubeEx}

\item People are detected in the current frame of the Master stream. There has been much work on human detection \cite{dalal2005histograms} and in particular on Pedestrian detection \cite{dollar2012pedestrian}.

\item Re-identification of people between the Master stream detections and the tubes extracted from the Slave streams is performed \cite{zheng2012transfer,Pedagadi_2013_CVPR,Zhao_2013_CVPR}. Re-identification scores between two objects are often given probabilistically e.g. \cite{Zhao_2013_CVPR}.

\end{enumerate}

\vspace{-0.1cm}
\subsection{Slave Action Tube Selection}
\label{subsec:opt}

In this section we assume a camera system consisting of one Master camera of real-time importance and one or more Slave cameras of forensic importance. We propose to detect people in the Master camera stream at fixed time intervals, and play for each Slave camera, the activity tubes from the past that contain the observed people. 

Pre-processing is done as described in Sec.~\ref{subsec:pre}. At fixed intervals of length $\delta T$ the Slave action tubes to be displayed in each Slave video $v$ are selected. The task is to select a set of tubes $S_v$ to display in Slave video $v$ out of the total set of tubes in the Slave view $B_v$ ($S_v \subseteq B_v$). There are three factors that are taken into account: i) displaying the maximal number of Slave tubes containing the people observed in the Master camera; ii) minimizing tube collisions; iii) a stable viewing experience: minimizing the number of tube switches at each interval.

This can be formulated using two energy terms, a collision term $E^C_v$ and an identical object overlap term $E^O_v$:
\begin{equation}
\label{eqn:master_tot}
E^T_v(S_v) = \alpha \cdot E^C_v(S_v) - E^O_v(S_v) ~~|~~ S_v \subseteq B_v
\end{equation}

We do not explicitly take into account the relations between different slave videos. The energy terms $E^O_v$ for Slave videos $v$ are optimized independently of other Slave videos.   

\subsubsection{Collision Cost}
\label{subsub:collision}

The objective of $E^C$ is to minimize collisions between action tubes placed in the generated videos. A small number of collisions can be tolerated, and it can greatly increase the number of Slave activities displayed simultaneously. The number of collisions that can be tolerated can be modified by adjusting $\alpha$ in Eq.~\ref{eqn:master_tot}.

Let tube $b$ be defined by binary function $\chi_b(x,y,t)$ indicating if the pixel $(x,y)$ in frame $t$ is active for tube $b$. 

Given a slave camera, the collision cost for its generated slave video $v$ is defined in Eq.~\ref{eq:tubecollision} (similar to \cite{pritch2008nonchronological}): the number of colliding pixels among all pairs of different tubes in the video. We add a discount factor for collisions that are forecast further away in the future as we become increasingly uncertain that the tubes will not be terminated before the forecast collision (due to new persons appearing and old persons disappearing in the Master view). The amount of discounting is determined by factor $d$.

\begin{equation}
\label{eq:tubecollision}
E^C_v(S_v) =  \sum_{b,\tilde{b} \in S^v} \sum_{x,y,t} \chi_b(x,y,t) \cdot \chi_{\tilde{b}}(x,y,t) \cdot d^t
\end{equation}
where $S_v$ is the set of tubes chosen for display in the output Slave video $v$ at the current time interval from the total set $B_v$.

\subsubsection{Identity Cost}
\label{subsub:ident}

The person identity cost in Eq.~\ref{eqn:master_overlap} encapsulates several requirements: i) displaying the Slave tubes having the highest probability of correspondence to the people detected in the Master stream (this set is labeled $O$). ii) making the number of tubes corresponding to each object in the Master frame roughly equal. iii) encouraging retention of already playing tubes for smoother viewing. This can be formulated as:
\begin{equation}
\label{eqn:master_overlap}
E^O_v(S_v) = \sum_{o \in O}   \sqrt{\sum_{b \in S_v} {(1 + \beta \cdot 1_{b \in S^{t-1}_v}) \cdot P_{b,o}}}
\end{equation}
Where $S^{t-1}_v$ is the set of Slave tubes selected in the last interval, and $\beta$ is a constant determining the strength of the preference to retain old tubes. The square root encourages the display of all objects in roughly equal numbers, otherwise most tubes may come from the same most likely object. When the Slave action tube and the person appearing in the Master camera are different persons this term has little effect, as the probability $P_{b,o}$ will be low.

\subsubsection{Cost Minimization}
\label{subsub:minimi}

The energy for each slave camera as expressed in Eq.~\ref{eqn:master_tot} can be minimized using standard discrete optimization methods. However the fast greedy approach described below generated good results as well.
\begin{itemize}[noitemsep,topsep=2pt,parsep=0pt,partopsep=0pt,leftmargin=*]
\item For all Slave videos $v$
\item Set $S_v = \phi$
\item Set list $L =  B_v$ (all tubes for video $v$)
\item Until no tubes left in $L$:
\begin{enumerate}[noitemsep,topsep=2pt,parsep=0pt,partopsep=0pt,leftmargin=*]
	\item For each tube $b \in L$ calculate the approximate decrease in overlap energy $\sum_{o \in O} \frac{(1 + \beta \cdot 1_{b \in S^{t-1}}) \cdot p^o_b}{ \sqrt{\sum_{\tilde{b} \in S_v} {(1 + \beta \cdot 1_{\tilde{b} \in S^{t-1}_v}) \cdot p^o_{\tilde{b}}}}}$  
	\item Select tube $b$ with the largest decrease.
	\item If the sum of collisions between $b$ and the tubes in $S_v$ is smaller than threshold $r$: if $\sum_{\tilde{b} \in S_v} \sum_{x,y,t} \chi_b(x,y,t) \cdot \chi_{\tilde{b}}(x,y,t) \cdot d^t < r$, add $b$ to $S_v$
		\item Remove $b$ from $L$.
\end{enumerate}
\end{itemize}
\
We use the binary update rule every $\delta T$ seconds and display the tubes $\left\{b|b \in S_v\right\}$ in slave view $v$.      

\vspace{-0.1cm}
\subsection{Synopsis Display of Slave Cameras}
\label{subsec:disp}

LVS is now generated for each slave camera $v$ by placing every tube from $S_v$ with the correct temporal offset (in case it has already been playing) over the stationary background of the corresponding Slave video. We emphasize that a synopsis video is created for each Slave camera. Tubes are not transferred between cameras, nor shifted in space. This ensure that all objects remain on their original background and geometries, creating videos that are easy to understand. 

\vspace{-0.1cm}
\section{Experiments}
\label{sec:exp}
\vspace{-0.1cm}

\begin{figure}[t]
	\centering
  \includegraphics[width=0.49\textwidth]{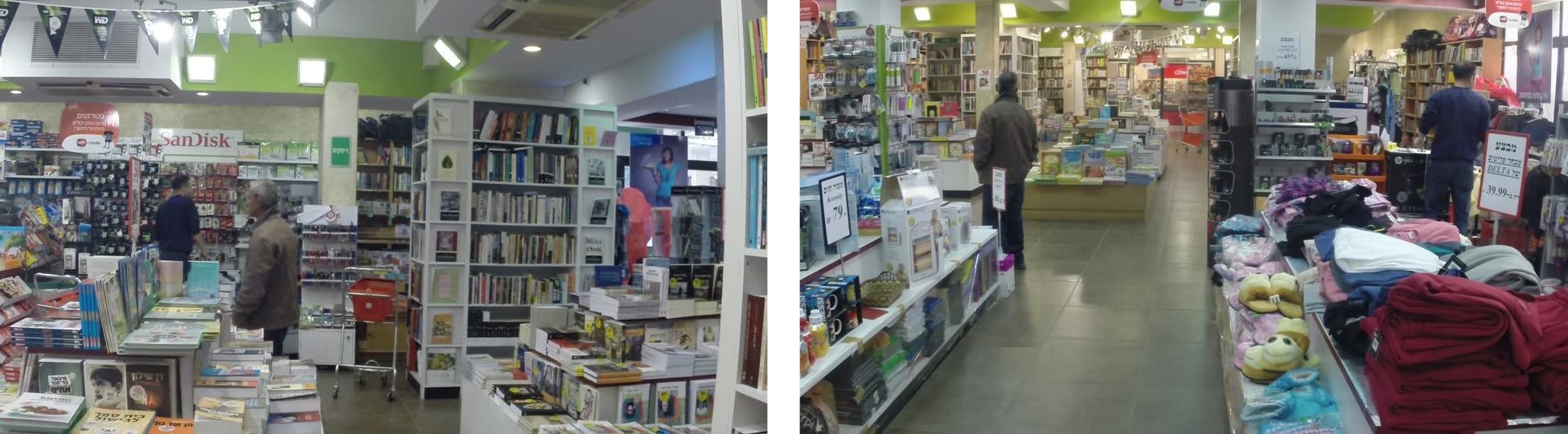}
	\caption{ A sample frame from the output of the MS algorithm on a Store scene. The left image is the Master camera and the right image is the Slave. Tubes corresponding to the two persons in the Master view were rendered simultaneously in the Slave videos. The clip can be seen at: http://www.vision.huji.ac.il/syncvid/}
	\label{fig:StoreFrame}
\end{figure}

\begin{figure}[t]
	\centering
  \includegraphics[width=0.49\textwidth]{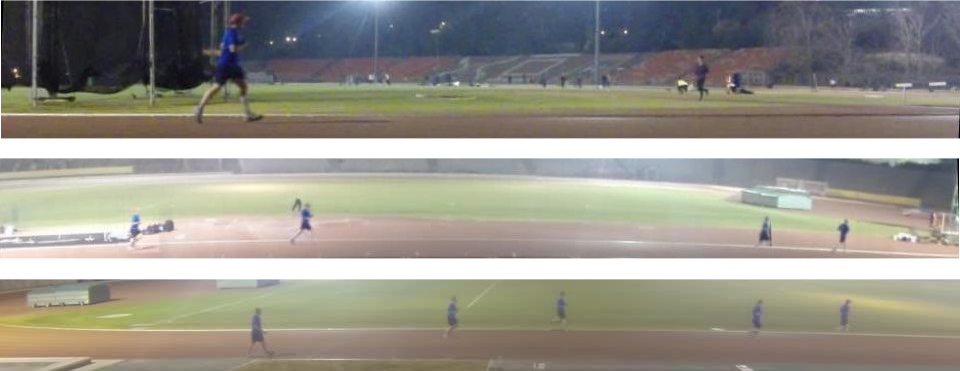}
	\caption{ A sample frame from the output of the Master-Slave algorithm run on a Stadium scene. The top video is the Master camera and the bottom two are Slaves. Multiple tubes corresponding to the person in the Master view are displayed in the slave videos. Many matching tubes were found and are displayed simultaneously. This cannot be achieved by frame-based methods. The clip can be seen at: http://www.vision.huji.ac.il/syncvid/}
	\label{fig:StadiumFrame}
\end{figure}

We present frames from two scenes, demonstrating the output of LVS. The Store Scene was recorded by two non-overlapping cameras in a store (Fig.~\ref{fig:StoreFrame}), the Stadium was recorded by three non-overlapping cameras around a stadium (Fig.~\ref{fig:StadiumFrame}). Tubes were extracted by state of the art background subtraction method such as \cite{Vibe}. Tubes were manually re-identified between Master and Slave tubes. Our method was then run using the following parameters: $\beta = 0.5$, $\delta T = 1~second$, $r=15$, $d = 0.978$. The output clips can be seen at http://www.vision.huji.ac.il/syncvid/.

\begin{figure}[t]
	\centering
  \includegraphics[width=0.23\textwidth]{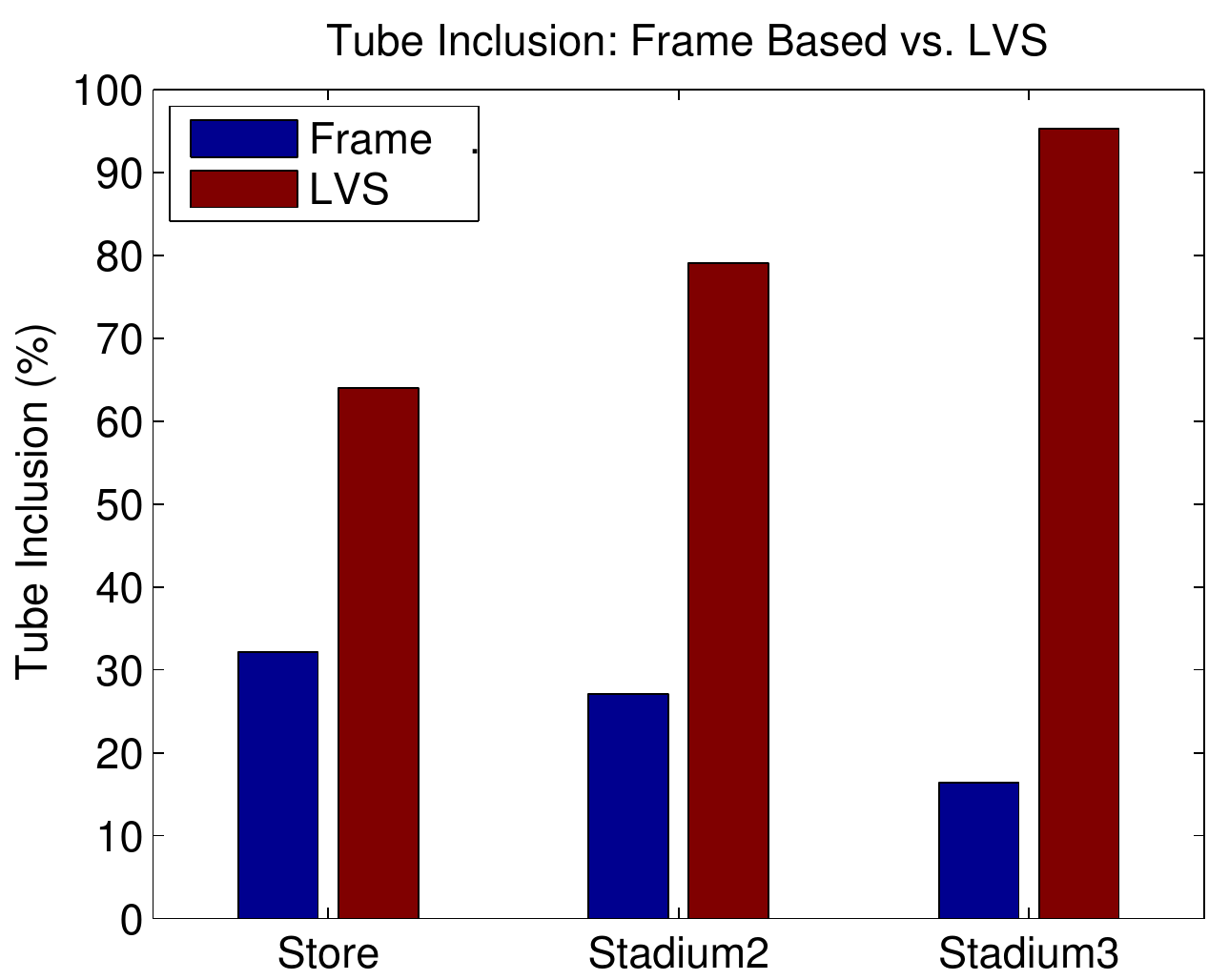}
  \includegraphics[width=0.23\textwidth]{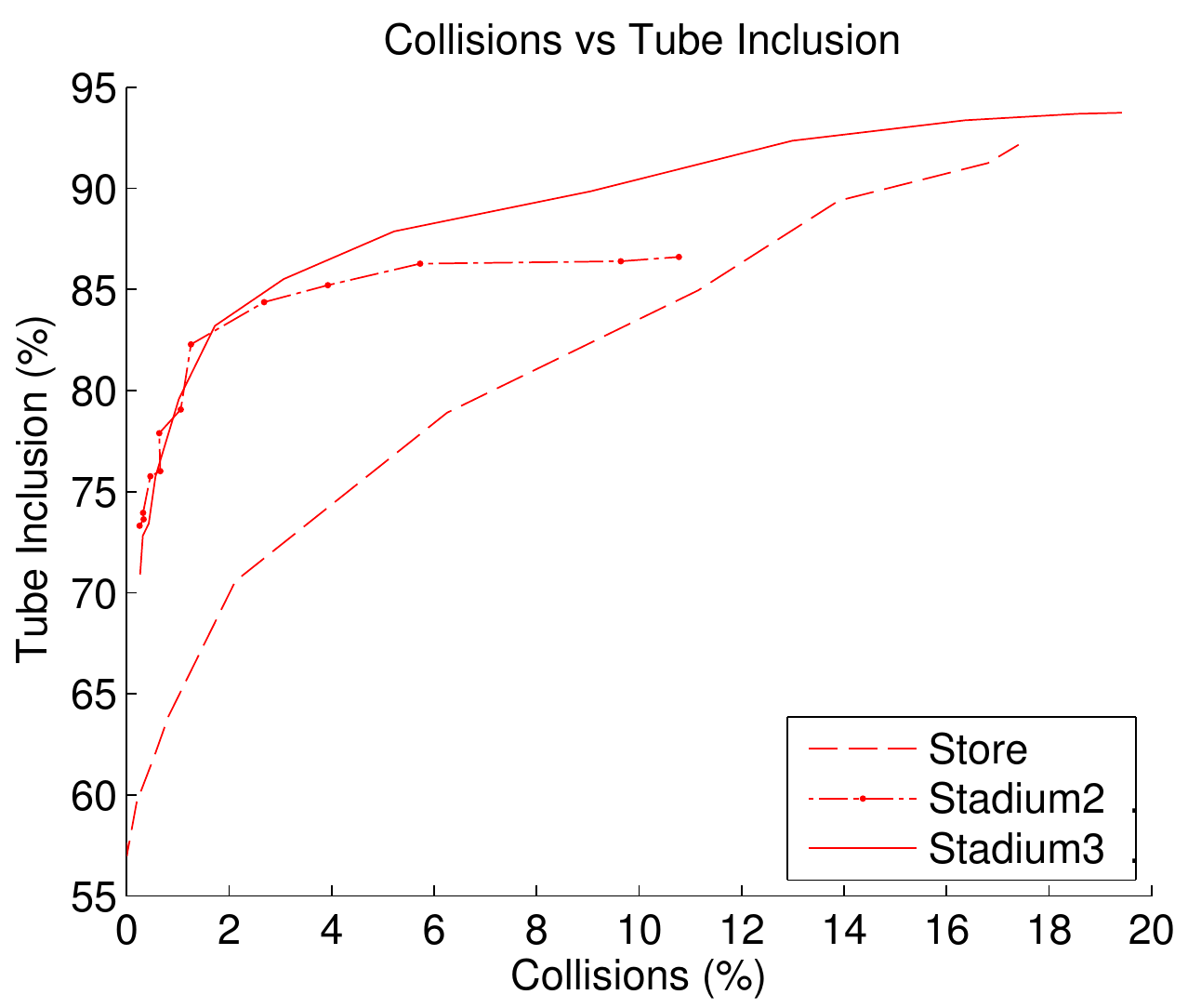}
	\caption{ a) Comparison of tube inclusion rates of LVS vs the frame based method for the three Slave videos. Significant improvements have been obtained. b) The Collision rate vs. the Tube Inclusion rate for the three Slave videos. 65-85\% of tubes can be included for a modest collision cost.}
	\label{fig:CollisionComp}
\end{figure}

Fig.~\ref{fig:CollisionComp}.a) shows a comparison between a frame-based method (showing the whole frames of the highest ranking Slave action tube) and our object-based method - LVS. The frame-based method was able to display only 15-30\% of the relevant Slave tubes, whereas our method was able to display 65-85\% of Slave tubes with minimal collisions. This performance-gap is expected to increase further when re-identification uncertainty is significant.

The trade-off between collisions and number of relevant Slave tubes can be seen in Fig.~\ref{fig:CollisionComp}.b) for the three Slave videos. A very modest collision rate (2\%) is required for displaying 65-85\% of relevant tubes.

Several benefits of LVS are apparent:

1) While concentrating on the Master camera, we are able to see much history of the objects in the Slave cameras. This can be of great utility for letting operators make decisions in real-time.

2) In many cases the Master stream is sparse and contains only a small number of objects, it is possible to display several candidate tubes for each object in the Slave nodes. This is helpful as in many scenarios, the top re-identification result has about 30\% recall probability, but the top 5 candidates have an accumulated recall probability of above 65\% \cite{Zhao_2013_CVPR}. Showing as many candidates as possible therefore increases the likelihood of seeing the whole history of the object across the scene.

\vspace{-0.3cm}

\section{Concluding Remarks}
\label{sec:conc}

\vspace{-0.2cm}

Live video synopsis is a novel object-based method for using summarization of previously recorded video for aiding live decision making. It was shown that our method has many advantages over frame based methods. Although in this paper we have concentrated on people, this method is general and can be used for any type of object that can be detected and re-identified across cameras (animals, cars, etc.). As our method relies on having a reliable object re-identification algorithm, improvements in person re-identification from video will increase the reliability of our method. More interestingly, our method can be used to display object re-identification examples for active learning algorithms. This can be used for obtaining interactive feedback from the operator for refining video re-identification performance.

\vspace{0.2cm}
    
\noindent\textbf{Acknowledgment:} This research was supported by Intel ICRI-CI, by the Israeli Ministry of Science, and by Israel Science Foundation.

\bibliographystyle{IEEEbib} 
\bibliography{egbib}

\begin{thebibliography}{10}

\bibitem{jiang2011anomalous}
Fan Jiang, Junsong Yuan, Sotirios~A Tsaftaris, and Aggelos~K Katsaggelos,
\newblock ``Anomalous video event detection using spatiotemporal context,''
\newblock {\em CVIU}, pp. 323--333, 2011.

\bibitem{zhao2011online}
Bin Zhao, Li~Fei-Fei, and Eric~P Xing,
\newblock ``Online detection of unusual events in videos via dynamic sparse
  coding,''
\newblock in {\em CVPR}, 2011.

\bibitem{kojima2002natural}
Atsuhiro Kojima, Takeshi Tamura, and Kunio Fukunaga,
\newblock ``Natural language description of human activities from video images
  based on concept hierarchy of actions,''
\newblock {\em IJCV}, pp. 171--184, 2002.

\bibitem{rohrbach13iccv}
Marcus Rohrbach, Wei Qiu, Ivan Titov, Stefan Thater, Manfred Pinkal, and Bernt
  Schiele,
\newblock ``Translating video content to natural language descriptions,''
\newblock in {\em ICCV}, 2013.

\bibitem{gong2000video}
Yihong Gong and Xin Liu,
\newblock ``Video summarization using singular value decomposition,''
\newblock in {\em CVPR}, 2000.

\bibitem{khoslalarge}
Aditya Khosla, Raffay Hamid, Chih-Jen Lin, and Neel Sundaresan,
\newblock ``Large-scale video summarization using web-image priors,''
\newblock in {\em CVPR}, 2013.

\bibitem{petrovic2005adaptive}
Nemanja Petrovic, Nebojsa Jojic, and Thomas~S Huang,
\newblock ``Adaptive video fast forward,''
\newblock {\em Multimedia Tools and Applications}, vol. 26, no. 3, pp.
  327--344, 2005.

\bibitem{rav2006making}
Alex Rav-Acha, Yael Pritch, and Shmuel Peleg,
\newblock ``Making a long video short: Dynamic video synopsis,''
\newblock in {\em CVPR}, 2006.

\bibitem{pritch2007webcam}
Yael Pritch, Alex Rav-Acha, Avital Gutman, and Shmuel Peleg,
\newblock ``Webcam synopsis: Peeking around the world,''
\newblock in {\em ICCV}, 2007.

\bibitem{pritch2008nonchronological}
Yael Pritch, Alex Rav-Acha, and Shmuel Peleg,
\newblock ``Nonchronological video synopsis and indexing,''
\newblock {\em IEEE-PAMI}, pp. 1971--1984, 2008.

\bibitem{feng2012online}
Shikun Feng, Zhen Lei, Dong Yi, and Stan~Z Li,
\newblock ``Online content-aware video condensation,''
\newblock in {\em CVPR}, 2012.

\bibitem{fu2010multi}
Yanwei Fu, Yanwen Guo, Yanshu Zhu, Feng Liu, Chuanming Song, and Zhi-Hua Zhou,
\newblock ``Multi-view video summarization,''
\newblock {\em IEEE Trans. Multimedia}, pp. 717--729, 2010.

\bibitem{sawhney2002video}
Harpreet~S Sawhney, Aydin Arpa, Rakesh Kumar, Supun Samarasekera, Manoj
  Aggarwal, Steve Hsu, David Nister, and K~Hanna,
\newblock ``Video flashlights: real time rendering of multiple videos for
  immersive model visualization,''
\newblock in {\em Proceedings of the 13th Eurographics workshop on Rendering},
  2002.

\bibitem{porikli2004multi}
Fatih Porikli,
\newblock ``Multi-camera surveillance: object-based summarization approach,''
\newblock {\em Mitsubishi Research TR-2003-145}, 2004.

\bibitem{dale2012multi}
Kevin Dale, Eli Shechtman, Shai Avidan, and Hanspeter Pfister,
\newblock ``Multi-video browsing and summarization,''
\newblock in {\em CVPRW}, 2012.

\bibitem{stauffer1999adaptive}
Chris Stauffer and W~Eric~L Grimson,
\newblock ``Adaptive background mixture models for real-time tracking,''
\newblock in {\em CVPR}, 1999.

\bibitem{ko2008background}
Teresa Ko, Stefano Soatto, and Deborah Estrin,
\newblock ``Background subtraction on distributions,''
\newblock in {\em ECCV'08}, 2008, pp. 276--289.

\bibitem{dalal2005histograms}
Navneet Dalal and Bill Triggs,
\newblock ``Histograms of oriented gradients for human detection,''
\newblock in {\em CVPR}, 2005.

\bibitem{dollar2012pedestrian}
Piotr Dollar, Christian Wojek, Bernt Schiele, and Pietro Perona,
\newblock ``Pedestrian detection: An evaluation of the state of the art,''
\newblock {\em PAMI}, 2012.

\bibitem{zheng2012transfer}
Wei-Shi Zheng, Shaogang Gong, and Tao Xiang,
\newblock ``Transfer re-identification: From person to set-based
  verification,''
\newblock in {\em CVPR}, 2012.

\bibitem{Pedagadi_2013_CVPR}
Sateesh Pedagadi, James Orwell, Sergio Velastin, and Boghos Boghossian,
\newblock ``Local fisher discriminant analysis for pedestrian
  re-identification,''
\newblock in {\em CVPR}, June 2013.

\bibitem{Zhao_2013_CVPR}
Rui Zhao, Wanli Ouyang, and Xiaogang Wang,
\newblock ``Unsupervised salience learning for person re-identification,''
\newblock in {\em CVPR}, June 2013.

\bibitem{Vibe}
M.~Van Droogenbroeck and O.~Paquot,
\newblock ``Background subtraction: Experiments and improvements for vibe,''
\newblock in {\em Change Detection Workshop at CVPR}, June 2012.

\end{thebibliography}

\end{document}